# Nish: A Novel Negative Stimulated Hybrid Activation Function

Yildiray Anagun[a*] and Sahin Isik[a]

[a]Department of Computer Engineering, Eskisehir Osmangazi University, Eskisehir, Turkey

*Corresponding author: yanagun@ogu.edu.tr

**Abstract**

An activation function has a significant impact on the efficiency and robustness of the neural networks. As an alternative, we evolved a cutting-edge non-monotonic activation function, Negative Stimulated Hybrid Activation Function (Nish). It acts as a Rectified Linear Unit (ReLU) function for the positive region and a sinus-sigmoidal function for the negative region. In other words, it incorporates a sigmoid and a sine function and gaining new dynamics over classical ReLU. We analyzed the consistency of the Nish for different combinations of essential networks and most common activation functions using on several most popular benchmarks. From the experimental results, we reported that the accuracy rates achieved by the Nish is slightly better than compared to the Mish in classification.

## 1. Introduction

ReLU (Krizhevsky et al., 2017; Nair and Hinton, 2010) has been used extensively as a non-linear function on deep neural networks; moreover it is one of the most preferred functions due to better convergence and gradient propagation compared to tanh, logistic and sigmoid. ReLU also suffers from the existence of certain three important cases such as non-zero mean, negative loss and unlimited output. Although, ReLU ensures training speed in the learning phase, since the derivative of the zero value region is also zero, learning does not occur in backpropagation (Dying ReLU problem). ReLU-n (Krizhevsky, 2012a) is another rectifier-based activation functions proposed to enhance the efficiency of the convolutional neural network (CNN). ReLU-n is a limited ReLU at *n*. When the value of *n* is chosen as 6 empirically, it is called ReLU6. This provides the model to learn the sparse input earlier. This is equivalent to assuming that each ReLU unit consists of only 6 replicated bias-shift Bernoulli units rather than an infinite amount (Lu et al., 2019). Leaky ReLU (LReLU) (Maas, 2013) tries to solve the vanishing gradient problem by multiplying by a constant $(a)$ for negative values of the function. Randomized Leaky ReLU (RReLU) (Xu et al., 2015) generates activation of neurons by random sampling for negative values. Parametric ReLU (PReLU) (Kaiming He et al., 2015a) is very similar to LReLU, but compared to it, the multiplier is a learnable parameter $(\rho)$. The equations of the ReLU and its variants are given as follows:

$$f_{ReLU}(x) = \begin{cases} 0 & \text{if } x < 0 \\ x & \text{if } x \geq 0 \end{cases} \qquad (1)$$

$$f_{LReLU}(x) = \begin{cases} ax & \text{if } x < 0 \\ x & \text{if } x \geq 0 \end{cases} \qquad (2)$$

$$f_{RReLU}(x)_{ij} = \begin{cases} a_{ij} x_{ij} & \text{if } x_{ij} < 0 \\ x_{ij} & \text{if } x_{ij} \geq 0 \end{cases}, \qquad (3)$$

$$a_{ij} \sim P(k,l), l > k \text{ and } k,l \in [0,1)$$

$$f_{PReLU}(x) = \begin{cases} \rho x & \text{if } x < 0 \\ x & \text{if } x \geq 0 \end{cases} \qquad (4)$$

Exponential Linear Unit (ELU) activation function (Clevert et al., 2016) gradually brings slope of the curve from a constant threshold to origin, in addition, it contains a negative saturation on the curve to manage variance of the forward propagating. It takes a binary value and is therefore often used as an output layer. Inspired by the ELU, Scaled ELU (SELU) (Klambauer et al., 2017) is its parametric activation function variant and does not have a vanishing gradient problem such as the ReLU. The SELU is defined for self-normalizing neural networks and deals with internal normalization. In other words, each layer preserves the mean and variance coming from the previous layers. It ensures this normalization by adjusting both the mean and variance so that the network converges faster. However, in deeper neural networks, it is possible to say that the SELU suffers from gradient explosion or loses its self-normalization property. Gaussian Error Linear Unit (GELU) (Hendrycks and Gimpel, 2016) performs element-wise neurons firing on a given input tensor. The GELU nonlinearity sets the weights of the inputs in according to the magnitude instead of their sign as in the ReLUs, and deals with the cumulative density function of the normally distributed inputs. Sigmoid Linear Unit (SiLU) (Elfwing et al., 2018) is calculated with the sigmoid function multiplied by its input, while the Swish activation function (Ramachandran et al., 2018) is obtained by adding a trainable *β* parameter, with slight modification to SiLU. Although all of these cutting-edge activation functions perform significantly better than previous classic activation functions when trained in 3 deep models, they converge more slowly compared to the ReLU, Leaky ReLU and PReLU. Based on the Swish's self-gating properties, a new self-regulating non-monotonic activation function Mish (Misra, 2020) tends to increase the efficiency of computer vision problems. Not only does the Mish ensure better empirical results than the Swish under most experimental conditions, but it also overcomes some of the Swish's drawbacks, particularly in large and complex architectures such as models with residual layers.

**2. Related Works**

Research on new heuristics/strategies that can outperform classical activation functions is nowadays one of the key points for finding solutions to image and signal processing problems with advanced deep learning mechanisms (Nwankpa et al., 2020). There is a close and complex relationship between the activation function and the CNN structure. Variants of the family of rectified units have been widely used in popular deep CNNs in recent years. (Clevert et al., 2015; Kaiming He et al., 2014; K. He et al., 2015b; Sermanet et al., 2014;

Zeiler and Fergus, 2014). After the activation function is defined for a specific deep learning model, activation is the same for neurons in all activation layers in that network. For instance, the flexible ReLU restated the adjusted point of the ReLU with a negatively shifted parameter (Loffe and Szegedy, 2015). (Hinton et al., 2012) developed an extension of dropout, which is called alpha dropout and is used with SELU. Thus, they provided distinct improvements in tasks such as speech and object recognition. To learn the best activation function, a different methodology toward genetic algorithm was suggested in (Basirat and Roth, 2018). Activation functions and normalization layers are key components in deep learning models and typically interact with each other. Another recent advance in this area is the evolution, which revealed the general structure of the function (Bingham et al., 2020; Liu et al., 2020). While gradient descent attempts to reach the global minimum during the training procedure, better performing custom activation functions can be designed and customized instead of the default activation for different architectures such as ResNet (Kaiming He et al., 2016a).

In deep learning models, hidden layers and activation functions are used to extract key features with nonlinear computations. In general, all neurons in a layer are passed through an activation function and a neuron's weight and bias values are determined according to the input value. In this way, the process of artificial neuronal firing occurs as communication between neurons. In neural science, lateral inhibition (LI) (Edwin Dickinson et al., 2022) is a mechanism by which excitatory neurons inhibit the activity of nearby neurons. Thus, it increases the contrast and sharpness in visual or audio sensory perception by reducing the activity of its neighboring neurons. If there are three neurons and one is the center, in this neural activity if the center neuron receives a stimuli, this is called the point of stimulation. However, we can see at this point the dendrites of surrounding neurons also get stimulated. Here, although they were less stimulated, LI were less occurred. In other words, LI involves the suppression of neurons by other neurons. For example, visual LI is the process in which photoreceptor cells improve edge perception while increasing the contrast in visual images. Considering that the inhibition of a neuron is essentially uncorrelated of the train dataset and variables, it is crucial for neural transmission to distinguish between informative and noisy signals, as large-scale datasets often contain several types of noise. We designed a novel activation function called the Negative Stimulated Hybrid Activation Function (Nish), which is adaptive to non-linear transformation and inputs as well as robust to noise.

The main achievements of this article to the literature are given in below.

1. We develop non-monotonic activation to combine the basic activation role in a data-driven way with a linear and non-linear perspective, respectively.
2. As an extension of the ReLU activation, it enhances a CNN architecture's ability to learn non-linear transforms and is designed to adapt to the inputs.

3. In order to objectively measure the results of the Nish, the default activation function used in modern deep CNN architectures are trained by replacing the proposed activation function. Thus, it provides performance gains in benchmarks at various scales, including robust and accurate learning in the optimization process.
4. In according to the loss and accuracy rates, the performance of the Nish shows its advantage over alternative recent activation functions in the literature such as SiLU (or Swish) and Mish.

The general flow of the article is structured as: the Nish function explained and formulated in Section 3. The compatibility of the Nish with the CNN architectures, experimental designs and general comparisons are referred in Section 4. Conclusions and future work are discussed in Section 5.

## 3. Definition of Nish

We will introduce the three latest activation functions (i.e. ReLU, SiLU and Mish) that determine how the outputs generated in the firing of neurons in deep neural networks should be modified, and then explain the design of the Nish activation function. Although, the ReLU provides a network run faster by assigning the value of zero in the negative regions, it brings the derivative to zero in case the value in the neuron is less than zero. For the sake of preventing of this problem, the SiLU and Mish perform learning of non-linear transformations for the negative axis. The SiLU is multiplied by a sigmoid function of an input value in the neuron and appears as a soft version of the ReLU, which is a continuous and undershooting version of the negative axis. The Mish requires a single scalar input (self-gating) similar to the Swish and was developed through methodical analysis and many features of the Swish such as an unbounded above, bounded below, non-monotonicity property, and continues derivative. Regarding the experiments of the Mish, under many conditions it gives better results than Swish and ReLU. The mathematical formula of SiLU is expressed as:

$$f_{SiLU}(x) = x * \sigma(x) = x * \frac{1}{1+e^{-x}} \quad (5)$$

where $\sigma(\cdot)$ is denoted as the logistic sigmoid wave and, the SiLU activation is nearly equal to the ReLU for x values in case of large magnitude.

Similar to the SiLU, continuous, self-regulating, non-monotonic Mish is mathematically formulated as

$$f_{Mish}(x) = x * \tanh(softplus(x)) \quad (6)$$

where $softplus(x) = \ln(1+e^x)$ and due to the prevent the loss of a small amount of negative data, it tries to eliminate the dying ReLU problem and reduce the activity of the neighbors of an excited neuron by increasing LI. To preserve these characteristics of a deep neural network, we propose Nish activation. Our aim is to improve learning accuracy and adaptation of the activation operation to the model by increasing slightly more convexity of the function on the negative axis. As shown in Fig. 1 (a), Nish is a new activation function with non-monotonicity property and we mathematically

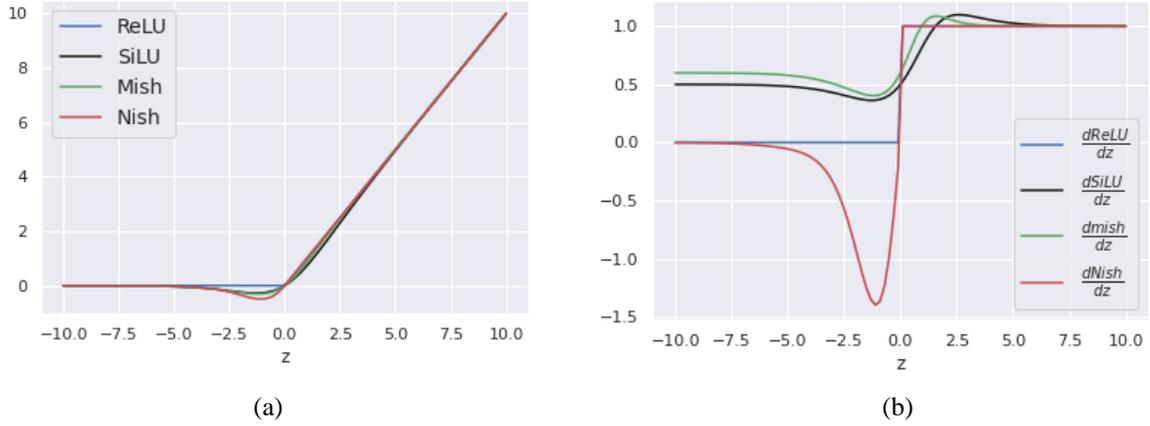

**Figure 1.** (a) The graph of activation functions the Nish, Mish, SiLU and ReLU. Unlike Mish and SiLU, Nish has slightly more convexity to preserve small weights in the negative axis. (b) The 1st derivative of selected activation functions.

formulated as follows:

$$f_{Nish}(x) = \begin{cases} x & \text{if } x \geq 0 \\ sigmoid(x)*(x+\sin(x)) & \text{if } x < 0, \end{cases} \quad (7)$$

Similar to both the SiLU and Mish, it is limited below domain and unlimited above domain with a range of [≈ -0.31, ∞). The 1st derivative of Nish, as shown in Fig. 1 (b), can be denoted as

$$\frac{\partial f_{Nish}(x)}{x} = \begin{cases} 1 & \text{if } x \geq 0 \\ x*(1-x)*(1+\cos(x)) & \text{if } x < 0, \end{cases} \quad (8)$$

Meanwhile, as mentioned before in the ReLU mathematical expression, neuron output is either equal to the input or to 0. Although this makes the training process faster, the hard gating property often results in some minor data deficiencies. Nish restricts the dying ReLU phenomenon because of the small amount of gradient being retained in the negative region, especially close to the zero value.

## 4. Experimental Analyses

This section explains the detailed and statistical experiments to assess the efficiency on the image classification tasks of the Nish activation function and to prove some improvement in the output results. By emphasizing the generalizability of the function, we first introduce the training methodology and different datasets, and then examine all of them in comparison with cutting-edge deep architectures, as well as statistical analysis and detailed evaluation.

### 4.1 MNIST Dataset

In the first experiment, we graphically show the test accuracy values for the ReLU, Swish, Mish, and Nish activation functions according to the number of layers using the MNIST dataset (LeCun et al., 2010), which has 70,000 samples in total (85% train and 15% test). The entire data set is in grayscale and each image is 28×28 pixels and categorized from 1 to 10. For this task, we used fully connected networks with linearly increasing depth, as implemented in Mish. The batch normalization process (Loffe and Szegedy, 2015) is utilized to decrease the internal covariate shift with 25% dropout (Srivastava et al., 2014). Also, the network is used Stochastic Gradient Descent (SGD) (Bottou, 2010) as optimizer and a 128 batch size. To make an equivalent comparison, a comparative study is performed between the proposed Nish and other activation functions, as shown in Fig. 2a. The stability of the network on test accuracy in case of using the Nish is better than that of using ReLU, Swish and Mish in the large models. Interestingly, after layer 23, a sharp drop in test accuracy can be seen for other activation functions, while Nish maintains stability of the accuracy. The other inherent shortcoming of activation functions is that they can fail since the inputs could be corrupted by random gaussian noise. The purpose of this operation is to determine the capacity of linear or non-linear weights of the relevant activation function to adapt the learning process to cope with noise. For this purpose, the entire MNIST data is degraded by adding white gaussian noise with gradually increasing standard deviation. Then, to analyze the activation functions sensitively, two experiments were conducted to show test losses, corresponding to 5-layered (Fig. 2b) and 10-layered (Fig. 2c) CNN architectures optimized using SGD, respectively.

As seen in Fig. 2b, the Swish, Mish and Nish functions show similar and better test losses compared to ReLU and proceed in a similarly against varying intensity of increasing input gaussian noise. In the contrast, Fig. 2c demonstrates that the activation functions fluctuate in test loss due to increased noise intensity in the deeper architecture. Especially, Nish acts as a more sensitive stimulus routine of neurons, cutting off the noise than the Swish, Mish and ReLU. At this point, we emphasize that the main scope of this study is to gain LI to the neural networks without compromising heavily on learning accuracy similar to the human nervous system.

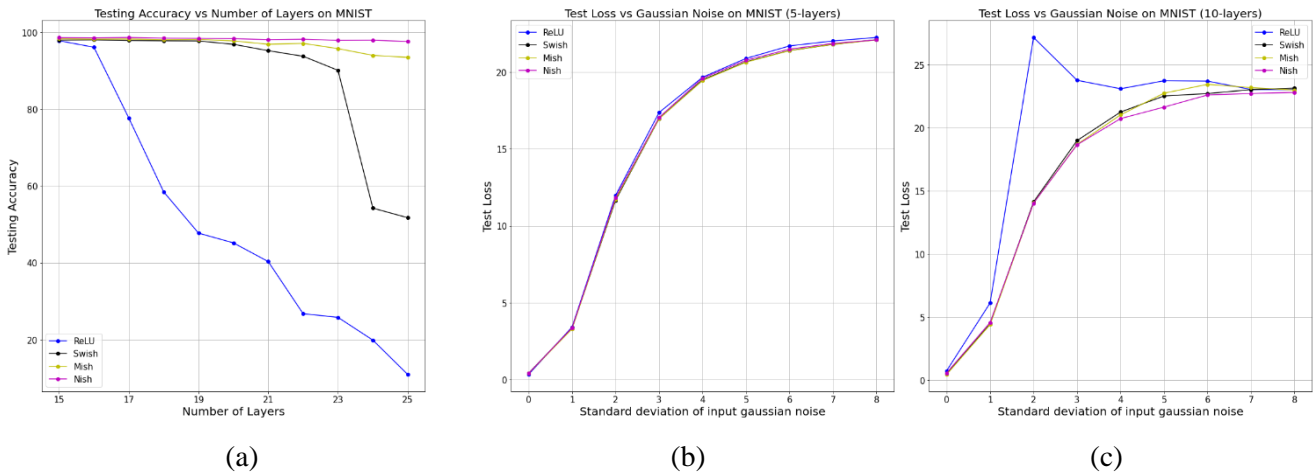

(a)  (b)  (c)

**Figure 2.** Comparison of Mish, Swish, ReLU and Nish (a) in terms of test accuracy with increasing depth of the neural network, (b) test loss with increasing input gaussian noise corresponding to 5-layered architecture and (c) test loss with increasing input gaussian noise corresponding to 10-layered architecture.

## 4.2 CIFAR Dataset

The second experiment was performed on the CIFAR-10 (Krizhevsky, 2012b) dataset containing 6,000 samples from each class and 10 labeled classes. In this experiment, the statistical significance and consistency improvement of Nish was evaluated and compared according to test loss, test accuracy and standard deviation for the classification task in SqueezeNet (Hu et al., 2018), ResNet18 (Kaiming He et al., 2016b), EfficientNetB0 (Tan and Le, 2019), EfficientNetv2s (Tan and Le, 2021) and MobileNetv2 (Sandler et al., 2018). To ensure standardization, the experiments conducted the same training setup that has 10 runs and each for 50, 100, and 150 epochs and the Adam optimizer (Kingma and Ba, 2014) is used. However, test accuracy verification of the Nish, only the corresponding activation function was changed while setting all hyper-parameters fixed. In Table 1, we first calculated final highest mean accuracy $\left(\mu_{acc}\right)$. Second, for further analyze lowest mean loss $\left(\mu_{loss}\right)$ and third lowest standard deviation of accuracy $\left(\sigma_{acc}\right)$ are calculated at each run compared with the top-specialized activation function Mish. As seen in the table, when we examine the difference between 50 and 150 epochs, SqueezeNet, ResNet18, and EfficientNetv2s require more cycles to achieve the greater level of $\mu_{acc}$. Since the value of $\mu_{loss}$ is very high for all 50 epoch cases, it does not establish for the concerned model very close to the learning behavior. We observe that the statistical results of the proposed activation function is accurate enough to provide approximately learning performance independent of the various models.

Upon inspecting the results, another important point is that the standard deviation $\sigma_{acc}$ of Nish at 150 epochs are less than Mish for almost all models, except for SqueezeNet. Thus, we showed further that the Nish is more robust than Mish when the model is close to convergence or fully converged. As the number of epoch increases, Nish activation continues to offer noteworthy better performance in terms of test accuracy than Mish.

## 4.3 ImageNet Dataset

To estimate the capability of the Nish, we should test it on unbalanced and substantial data sets. This is essential in order for us to arrive at inferences about its reliability and its contribution. In order to accomplish this goal, analyses were conducted using a variety of activation functions on the ImageNet-100 (Deng et al., 2009) dataset, which was chosen since it is both extensive and imbalanced. Our hypothesis is that the Nish activation function can deliver good performance over multi-class and huge datasets. In this study, it was underlined that activation methods such as classical ReLU disregard the information close to zero. To tackle this problem, a combination of sinus-sigmoidal approaches was used by preserving the discriminative information. This research aims to highlight the fact that negative values relatively near zero carry meaningful information after neurons have been triggered. We generated the Top-1 accuracy for both the Nish and the Mish as given in Table 2. We have used only two augmentation Random Horizontal Flip and Random Vertical Flip

**Table 1.** Statistical comparison of Mish and Nish activation functions in an image classification task on the CIFAR-10 dataset using various models at 50, 100, and 150 epochs and 10 runs.

| Model Name | | Mish | | | Nish | | |
|---|---|---|---|---|---|---|---|
| | | 50 | 100 | 150 | 50 | 100 | 150 |
| SqueezeNet (Hu et al., 2018) | $\mu_{acc}$ | 86.07% | 90.98% | 91.36% | **86.17%** | **91.06%** | **91.51%** |
| | $\mu_{loss}$ | 0.4204 | **0.2738** | 0.3448 | **0.4189** | 0.2771 | **0.3420** |
| | $\sigma_{acc}$ | 0.6281 | 0.2201 | **0.2567** | 0.4730 | 0.1293 | 0.2696 |
| ResNet18 (Kaiming He et al., 2016b) | $\mu_{acc}$ | 85.98% | 90.93% | 91.22% | **86.01%** | **91.04%** | **91.38%** |
| | $\mu_{loss}$ | **0.4288** | 0.2738 | **0.3296** | 0.4323 | **0.2731** | 0.3343 |
| | $\sigma_{acc}$ | 0.8765 | **0.1638** | 0.2905 | **0.5970** | 0.1844 | **0.1496** |
| EfficientNetB0 (Tan and Le, 2019) | $\mu_{acc}$ | 85.25% | 88.13% | 88.45% | **85.59%** | **88.55%** | **88.57%** |
| | $\mu_{loss}$ | 0.4373 | 0.3538 | **0.3818** | 0.4285 | 0.3452 | 0.3850 |
| | $\sigma_{acc}$ | **0.2374** | 0.2085 | 0.2783 | 0.2651 | **0.1177** | **0.2255** |
| EfficientNetv2s (Tan and Le, 2021) | $\mu_{acc}$ | **85.39%** | 90.31% | 90.67% | 84.10% | **90.34%** | **90.76%** |
| | $\mu_{loss}$ | **0.4435** | **0.2934** | **0.3489** | 0.5015 | 0.2987 | 0.3727 |
| | $\sigma_{acc}$ | 1.2926 | 0.2807 | 0.2183 | **0.6429** | **0.0057** | **0.2054** |
| MobileNetv2 (Sandler et al., 2018) | $\mu_{acc}$ | **86.06%** | 89.06% | 89.59% | 86.01% | **89.33%** | **89.64%** |
| | $\mu_{loss}$ | **0.4115** | 0.3227 | **0.3430** | 0.4162 | **0.3201** | 0.3491 |
| | $\sigma_{acc}$ | 0.4694 | **0.0945** | 0.1693 | 0.3737 | 0.1755 | **0.0996** |

with 0.5 probability rate. For all models batch size is selected 32.

Although, it was seen that the Mish approach performs comparatively worse than the Nish in the absence of pretrained weights, however, one can say that that the Mish method is more effective when there are weights pretrained on ImageNet-1k with different activation functions. It demonstrates that Mish has limitations regarding embedding for developing a model without pretrained CNN models.

**Table 2.** Top1 scores for ImageNet-100.

| Metric | Augmentation | Pretrained | Mish | Nish |
|---|---|---|---|---|
| mobilenetv2 (Sandler et al., 2018) | yes | no | 46.17% | **46.54%** |
| tinynet_c (Chen et al., 2019) | yes | no | 66.69% | **66.77%** |
| effnetb0_ap (Wightman, 2009) | yes | yes | **80.96%** | 80.64% |
| tinynet_b (Chen et al., 2019) | yes | yes | **83.98%** | 83.92% |

## 5. Conclusions

Because of the presence of multiple nested layers in CNN deep networks, repeated growth in output values of neurons causes the gradient to become progressively smaller in forward propagation and "vanishing" of weights, which saturates the network in back propagation. Therefore, considering activations in deep learning algorithms for picking a stable solution is very useful. To effectively handle this problem, which is encountered on the bases of gradients in both shallow and deep architectures, a new analytics activation function capable of providing LI must be developed and applied to fulfill predictions in the negative region with reasonable accuracy. This study proposes a new activation function is called Nish, in connection with an image classification task. Therefore, the main focus lies on activation functions that are better suited for image classification than the default activation functions of recent deep network models. After giving examples of some classical ReLU-based activations, an introduction of the most frequently used activation functions such as Mish and Swish is given and the general formulation of Nish is presented. The deep learning architectures and their compatibility with different datasets were also examined. It turned out that the Nish yielded the best empirical performance among all other activations such as the ReLU, Swish and Mish for image classification. An advantage of the Nish is its ability to better converge at high epochs count in deep architectures such as EfficientNetB0 and EfficientNetv2s. This is important since, if the presence of noise, a model can be overfitted and the model does not make accurate predictions on testing data. Nish has proved to be more robust under noisy circumstances, and the performance degrades gracefully compared to the ReLU, Swish and Mish when the noise intensity increases. As discussed in before, there are still many possibilities for future improvement. For example, optimizing Nish reduces the computational cost and speeds up the training time, in addition, evaluating performance of the Nish in various image processing tasks and other recent models in regression. We also encourage that our results motivate developers of future deep network models to provide an appropriate training mechanism that could be used to implement layers.

**Acknowledgments**

This research did not receive any funding support from public or private sectors.